\documentclass[10pt, conference]{IEEEtran}

\usepackage{amsmath,amssymb,amsfonts}
\usepackage{graphicx}
\usepackage{pifont}

\usepackage{amsthm}
\usepackage{xcolor}
\usepackage[figuresright]{rotating}

\usepackage{graphicx}
\graphicspath{{/}{fig/}}

\usepackage{array}
\usepackage{textcomp}
\usepackage{xcolor}
\usepackage{multirow}
\usepackage{booktabs}

\usepackage{mathtools}
\usepackage{breqn}
\usepackage{float}

\usepackage{pgfplots}
\pgfplotsset{compat=1.7}
\usepgfplotslibrary{groupplots}

\usepackage{caption}
\usepackage{subcaption}

\usepackage{multirow,tabularx}
\usepackage{hyperref}
\usepackage{flushend}
\usepackage{algorithmic}
\usepackage[vlined, ruled, shortend]{algorithm2e}

\usepackage{soul}

\newlength\figureheight
\newlength\figurewidth
\SetAlCapNameFnt{\footnotesize}
\SetAlCapFnt{\footnotesize}

\title{
    Secure Heterogeneous Multi-Robot Collaboration and Docking with Hyperledger Fabric Blockchain \\
}

\author{
    \IEEEauthorblockN{
        \vspace{1em}
        Salma Salimi\IEEEauthorrefmark{2},
        Paola Torrico Mor\'on\IEEEauthorrefmark{2},
        Jorge Pe\~na Queralta\IEEEauthorrefmark{2},
        Tomi Westerlund\IEEEauthorrefmark{2}
    }
    \IEEEauthorblockA{
        \normalsize
        \IEEEauthorrefmark{2}\href{https://tiers.utu.fi}{Turku Intelligent Embedded and Robotic Systems (TIERS) Lab, University of Turku, Finland}.\\
        Emails: \textsuperscript{1}\{salmas, pctomo, jopequ, tovewe\}@utu.fi\\[+6pt]
    }
}

\begin{document}

\maketitle
\thispagestyle{empty}
\pagestyle{empty}



\begin{abstract}%
    \label{sec:abstract}%
    In recent years, multi-robot systems have received increasing attention from both industry and academia. Besides the need of accurate and robust estimation of relative localization, security and trust in the system are essential to enable wider adoption. In this paper, we propose a framework using Hyperledger Fabric for multi-robot collaboration in industrial applications. We rely on blockchain identities for the interaction of ground and aerial robots, and use smart contracts for collaborative decision making. The use of ultra-wideband (UWB) localization for both autonomous navigation and robot collaboration extends our previous work in Fabric-based fleet management. We focus on an inventory management application which uses a ground robot and an aerial robot to inspect a warehouse-like environment and store information about the found objects in the blockchain. We measure the impact of adding the blockchain layer, analyze the transaction commit latency and compare the resource utilization of blockchain-related processes to the already running data processing modules.
\end{abstract}

\IEEEpeerreviewmaketitle



\section{Introduction}\label{sec:introduction}

Multi-robot systems (MRS), including aerial and ground robots, are used more extensively in recent years, gaining robustness and adaptability for their missions~\cite{rizk2019cooperative}. By utilizing heterogeneous multi-robot systems, a wide range of tasks with different characteristics can be autonomously and simultaneously accomplished in the same environment~\cite{queralta2020collaborative}. To make this happen, we need seamlessly integrate multiple technologies. Integrating multiple technologies, security issues must be considered carefully as they are an intrinsic part of multi-disciplinary field of robotics~\cite{aditya2021survey}.

\begin{figure}
    \centering
    \includegraphics[width=.49\textwidth]{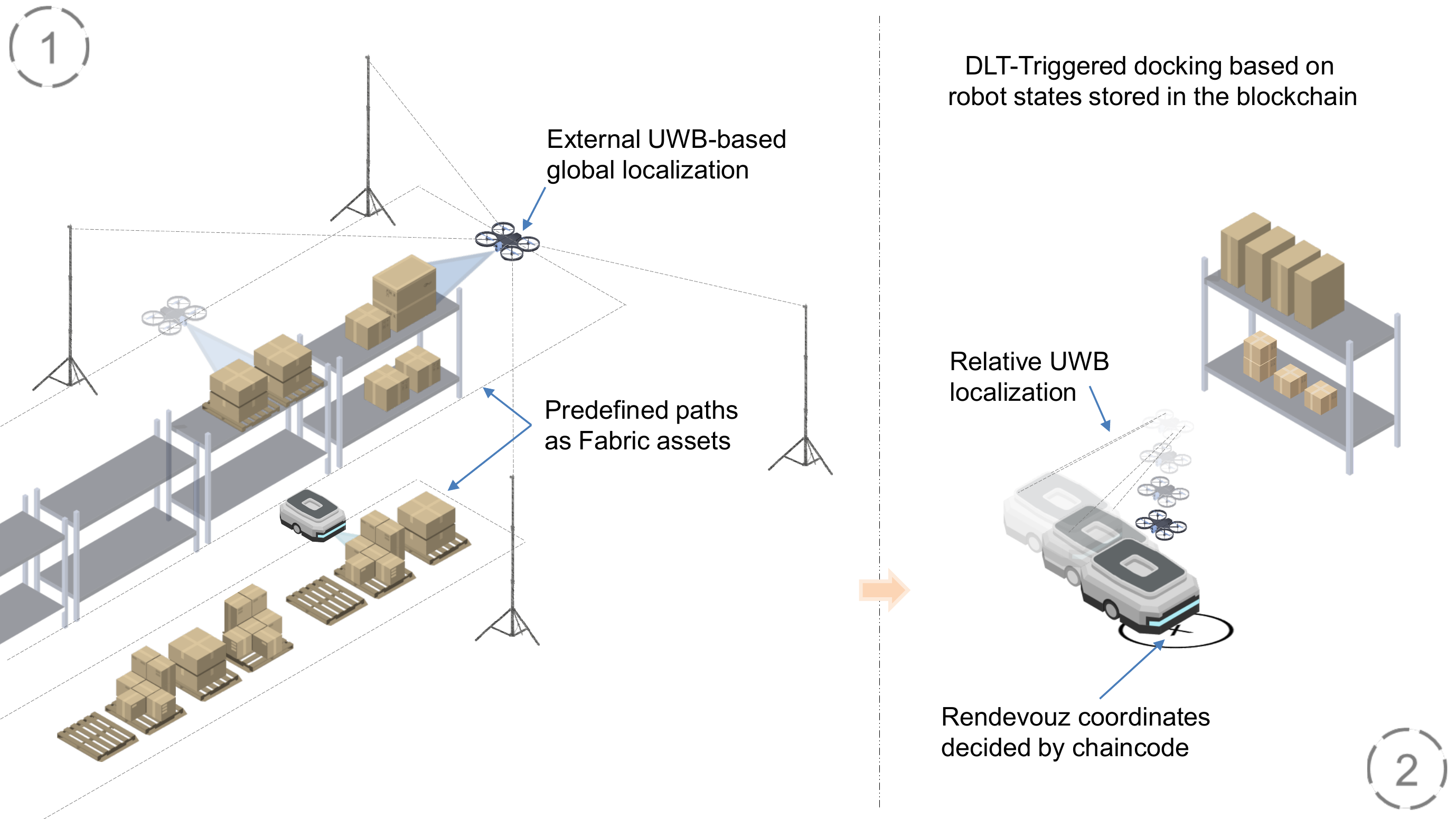}
    \caption{Conceptual illustration of the proposed application scenario. The blockchain is used for high-level mission commands (predefined robot paths), for triggering multi-robot cooperative actions based on chain-recorded robot states (docking) and switching operation modes (activate/deactivate global/relative localization UWB beacons). Detected objects are also stored in the blockchain.}
    \label{fig:concept}
\end{figure}

Blockchain technology is enabling more secure and trustable distributed systems that multi-robot systems are in essence. Permissioned or private blockchains have special potential in industrial applications, allowing for transparent cooperation between organizations while ensuring data remains private. This is also an important consideration when managing large-scale systems of autonomous robots, which may need to cooperate with each other but also produce data that should remain private to a subset of parties. In this paper, we explore the potential of the Hyperledger Fabric blockchain for managing cooperation between robots in addition to managing individual autonomous robots in a fleet.

In distributed systems, blockchain technology can provide security~\cite{qian2018towards}, trust~\cite{song2018blockchain}, data management~\cite{ayoade2018decentralized}, peer-to-peer transactions and a fault-tolerant middleware~\cite{chen2018devify}. Based on how they handle user credentials, blockchain platforms can be classified into two main categories: (i) public or permissionless, and (ii) private, permissioned, or consortium blockchains~\cite{queralta2021blockchain}. However, we are particularly interested in private blockchains for managing private robot fleets and maintaining data produced by robots, or part of it, private.

Hyperledger Fabric is a permissioned blockchain. It has a modular architecture supporting plug-in components, and it also achieves data isolation via channels so that two or more peers can have private communication in the network~\cite{aditya2021survey}, \cite{queralta2020enhancing}.

Due to the increased potential for multi-modal and multi-source sensor fusion in multi-robot scenarios, and the flexibility and robust nature of deployment in complex environments, heterogeneous multi-robot systems and algorithms for collaborative autonomy are also getting increasing interest~\cite{queralta2020end}.

Multi-robot systems, where each robot can do a particular task, can contain heterogeneous robots that communicate and coordinate in order to accomplish a particular task. Multi-robot systems should be able to eliminate false information presented through byzantine entities and/or resist hacking and manipulation of data and path planning systems that could result in disastrous outcomes. Blockchains provide a decentralized, secure and trusted platform that enables coordinating and allocating tasks, path-planning and controlling processes~\cite{afanasyev2019towards}.

Hyperledger Fabric is a private permissioned blockchain framework developed in Go. With highly modular, configurable, and pluggable features, developers are free to implement the technologies of their choice, and to do so in whatever way they want to (e.g., consensus protocols)\cite{alvares2021blockchain}.

Collaboration between different ground or aerial vehicles is necessary to accomplish a mission. It is particularly essential that unmanned aerial vehicles (UAV) and unmanned ground vehicles (UGV) coordinate during missions in remote areas where humans are in potentially hazardous situations\cite{hussein2021key}. 
Robots are capable of performing these tasks safely, as they can collect data from the environment easily, and also perform preprogrammed tasks by seeking around in certain scenarios. It is an interesting challenge to give them autonomous decision and also real-time cooperation and adaption with minimum intervention of human.

Ultra-Wideband (UWB) is a radio frequency technology that is used for indoor localization. Since these devices are more affordable than other indoor localization solutions, UWB technology is attracting increasing attention~\cite{ridolfi2021self}. Anchors and tags are the two types of nodes used in UWB localization. To obtain the position of tags in most anchor-based scenarios, the locations of the anchors must be known. The coordinates of the fixed anchor nodes are assumed to be known by the system for this calculation. In the presence of enough fixed nodes, depending on the ranging method implemented, the tags can estimate their distances from each anchor and calculate their own location using algorithms such as multilateration~\cite{ridolfi2018analysis}.

In summary, this paper extends our previous works in blockchain-based fleet management with Hyperledger Fabric~\cite{salimi2022towards}, leveraging the blockchain identities for collaboration between ground and aerial robots. Specifically, we utilize Fabric smart contracts for logging historical data, and collaborative decision-making. We put this approach into practice with an inventory management application where a ground and an aerial robot inspect a warehouse-like environment and store information about objects found in the blockchain. The smart contracts are then also used to enable docking of the aerial robot in the ground robot when the battery is too low. The robots use anchor-based UWB localization for following predefined trajectory, while the docking smart contract triggers the activation of anchors in the ground robot for more accurate relative localization while docking. These smart contracts can be extended for other collaborative tasks.

The rest of this manuscript is organized as follows. In Section II we introduce related works in blockchain solutions for robot fleets and UWB-based localization for multi-robot systems in GNSS-denied environments. Section III describes the system design, with Section IV focusing on the experimental settings and methodology. Section V reports experimental results and Section VI concludes the work.


\section{Background} \label{sec:related_work}

This section reviews the literature in the areas of blockchain technologies for multi-robot systems first, and then delves into specific applications to aerial and ground robots.

Considering time limitations, execution and autonomy requirements in multi-robot path planning approaches, authors in~\cite{mokhtar2019blockchain} present a distributed control system which is time-sensitive and is using the permissioned and private blockchain, Hyperledger Fabric platform. As a result, Hyperledger Fabric, which has been used in this paper, reveals less transactional latencies compared to permissionless, public blockchain platforms. It is  also more flexible and still features distributed execution of logic that enforces the consensus attainment for the collaborative members by the use of smart contracts. From the perspective of building trust, and to provide a trusted edge collaborative inference environment, the work in~\cite{li2021blockchain} introduced a blockchain-based collaborative edge knowledge inference framework, which ensures the reliance of data sharing, avoids knowledge pollution and detects malicious nodes.

A challenge in collaborative multi-robot systems is organizing the communication in data exchanges.
To optimize the amount and type of data on the exchange between them, a novel approach for managing the terms of these systems with blockchain technology has been illustrated by~\cite{queralta2019blockchain}. The proof of work systems for estimating the availability of computational resources of different robots have been proposed in this paper, and also smart contracts have been integrated to rank and analyze the quality and validity of each of the robots sensor data. 

In a similar direction in terms of managing data transmissions, Guo et al. proposed the integration of edge computing on a blockchain framework to enable large amounts of data excjanges in a spherical multi-robot system\cite{guo2020study}, instead of overloading each robot's nodes. Additionally, the authors recommended a distributed data processing system that exploits blockchain and edge computing technologies, in order to solve byzantine fault-tolerance problems for entities with limited resources. These works lay the background for more complex, real-world deployments as we are introducing in this paper.

A key use case of blockchain technology in multi-robot systems is also role allocation and driving the collaboration between the robots. Blockchain-based solutions have been shown to increment interaction efficiency, providingnmore reliable data exchange, consensus in trustless situations, detecting performance issues, identifying intruders, assigning tasks, or deploying distributed solutions and joint missions~\cite{afanasyev2019towards}. This overview of applications shows the potential of blockchain technologies, but also identifies limitations of current solutions that we are addressing in this paper with a permissioned blockchain platform.

In another recent application, Castell\'o Ferrer et al. use a blockchain as an asynchronous registry of messages by leaders communicating with robots with gregarious attributes to obtain security, limited memory, and resilience to deception~\cite{ferrer2021following},. The authors present solutions for the Byzantine Follow Multiple Leaders and Byzantine Loosely Follow Multiple Leaders problems where it has been improved that the proposed method could be utilized in practical scenarios where there are resource limitations. We also study in this paper the impact in terms of resource utilization of the blockchain layer, which we show to be negligible in complex robotic systems.

In terms of blockchain applications to drones, UAVs are a type of aerial vehicle that in addition to being able to operate autonomously, also possess the ability to fly with preplanned flight missions or create their own plan mid-flight~\cite{alvares2021blockchain}. Blockchain technology has been used for protecting the privacy and security of the entire trade process in~\cite{xu2021edge}, where a Stackelberg dynamic game based trading scheme has been presented to solve the edge computing resource allocation issue between Edge Computing Stations (ECSs) and Unmanned Aerial Vehicles (UAVs).

The applications to UAVs are multiple and varied. In a relevant work more related to industrial use cases, and according to the high demand of decentralized solutions for enforcing regulations, satisfying cybersecurity requirements and a need for efficient air traffic management, Alkadi et al. have described several issues with the current UAV traffic management systems and presented a solution that relies on cooperation between the concepts of blockchain smart contracts and mobile crowdsensing~\cite{alkadi2021unmanned} .

From the perspective of multi-robot system deployments, Aloaily et.al. introduced a network solution that fulfills the capabilities of UAVs and UGVs, as well as improving connectivity, service availability, and energy efficiency \cite{aloqaily2021energy}. Additionally, the solution employs a service composition strategy to deliver user-specific services based on their requirements, and also a consensus algorithm is used to verify locally trained models with blockchain to support integrity and authenticity of sensitive services.
Compared to previous works, in this paper we have used Fabric smart contracts for collaboration decision-making while logging historical data and leveraged the blockchain identities for multi-robot collaboration by presenting an inventory management application.  

Finally, regarding practical considerations for the deployment of autonomous robots in GNSS-denied environments, there is a number of works in the literature that leverage UWB localization for inventory management and other indoor operations in warehouses~\cite{xianjia2021applications, macoir2019uwb, xianjia2021cooperative}. These are just a few examples of wider use of UWB technology in GNSS-denied environments~\cite{xianjia2021applications}. Relative localization between ground and aerial robots enabled by UWB ranging makes possible more efficient collaboration~\cite{queralta2020vio}. In comparison to other approaches of cooperative relative localization, wireless ranging provides high performance and low complexity of the system. UWB in particular offers unparalleled performance in unlicensed bands, resilience to multi-path, low interference with other radio technologies and also high time resolution where in this paper we switch between external localization for navigation and relative localization for autonomous docking of the aerial robots on the ground robots~\cite{shule2020uwb}.


\begin{figure*}
    \centering
    \includegraphics[width=\textwidth]{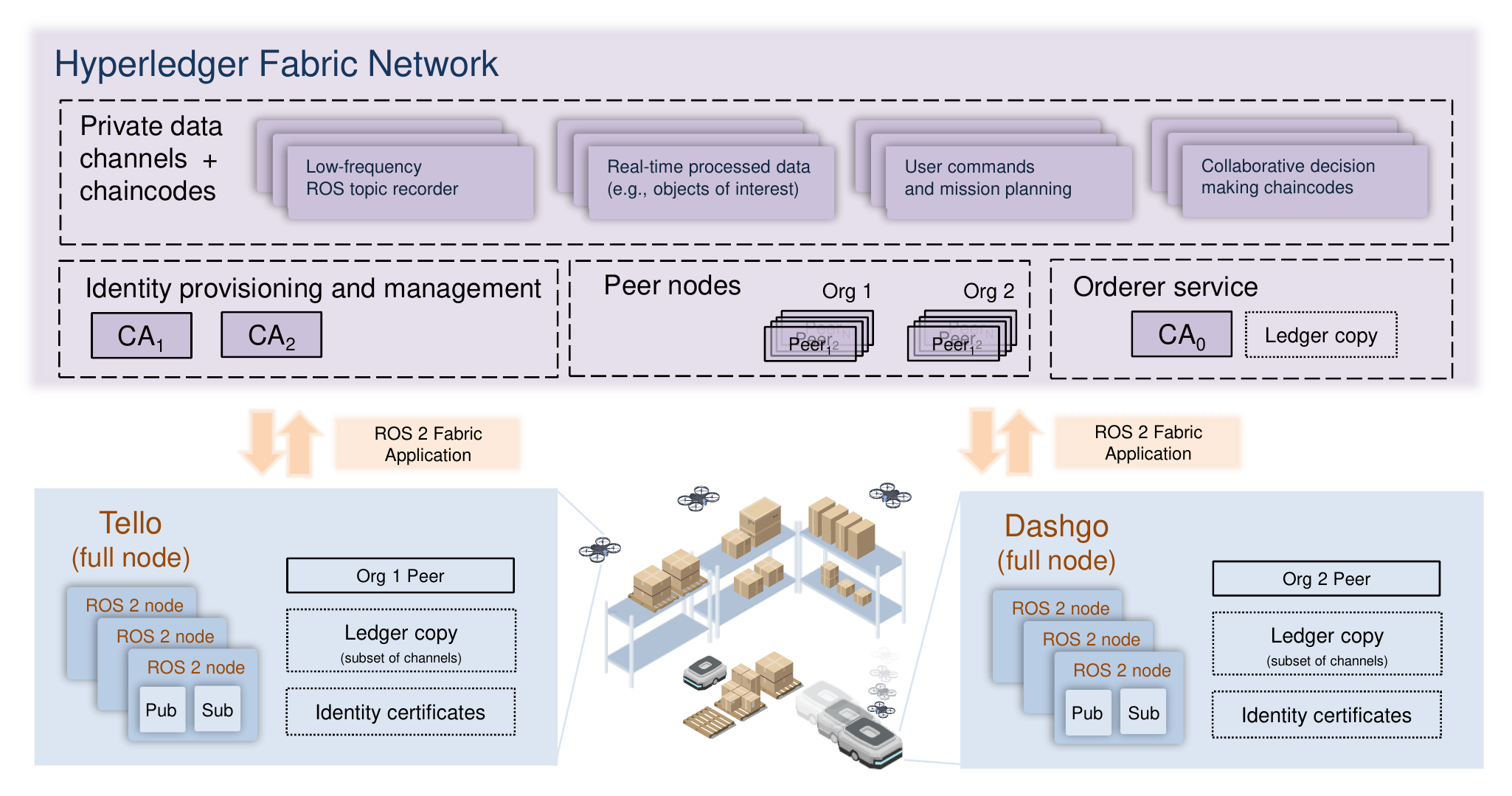}
    \caption{Architectural diagram of the proposed framework.}
    \label{fig:framework_architecture}
\end{figure*}

\section{System overview}

The growing interest in multi-robot systems is driven, in part, by the complexity of tasks and performance requirements. Multiple applications exist in a variety of scenarios, from emergency and rescue missions to surveillance applications and collaborative tracking~\cite{queralta2020collaborative, madridano2021trajectory}. Integrating DLTs into such systems can benefit identity management, data sharing security, monitoring and multi-robot inventory management and consensus. Knowledge-sharing-based collaborative inference is often necessary in multi-robot systems to accomplish complex tasks. In order to protect the integrity of such data sharing, it is essential to establish a secure environment. With DLTs, security and privacy can be enhanced while ensuring preservation of data integrity.

\subsection{Fleet management with Hyperledger Fabric}

Hyperledger Fabric is private and permissioned blockchain which breaks it from other blockchain systems. Instead of a public and permissionless system that allows anyone to join network, Fabric members join through a trusted Member Service Provider (MSP). In industrial robot fleets, Hyperledger Fabric blockchains can be used to manage identity, secure control interfaces, auditable data flows and create private data channels~\cite{salimi2022towards}. Therefore, the main goal of this paper is cooperating of multi-robot systems for a specific purpose using Hyperledger Fabric network for identity management of robots in a fleet, secure data management and robots control. On the other hand, Fabric smart contracts has been used in this paper for storing path and mission parameters or collaboration decision making.

\subsection{Localization and deployment}

Autonomous operation in GNSS-denied environments can be achieved with a variety of onboard and external sensors or landmarks. In recent years, UWB positioning systems have become increasingly popular owing to their relative high accuracy at low price~\cite{xianjia2021applications}. Autonomous robots can operate based on either external, global positioning~\cite{queralta2020uwb}, or relative positioning within multi-robot systems~\cite{queralta2020vio}. In this paper, we combine both localization approaches utilizing the blockchain as a channel for deciding when to switch between one mode or another.

In general terms, the core idea of the system we propose is that different modes of operation can be controlled via the Fabric smart contracts. For instance, robots can switch from individual autonomous operation to a behaviour involving collaboration. In our use-case scenario, this is exemplified by changing between external and relative localization methods. However, we present a generic architecture that can be adjusted to meet the needs of other application scenarios, from distribution of computation to role allocation.

\subsection{Architecture}

The proposed system architecture is illustrated in Fig.~\ref{fig:framework_architecture}. We use the Fabric to ROS~2 connection introduced in~\cite{salimi2022towards} to connect the ROS~2 nodes running on the robots to the Fabric blockchain. This is done by implementing ROS~2 nodes in Go using \textit{rclgo}\footnote{\url{https://github.com/tiiuae/rclgo}}. Compared to our previous work, we delve into a more complex use case with higher number of chaincodes in this paper, representing a more realistic industrial application. The blockchain layer sits conceptually on top of the data processing and robot control software, driving the overall mission parameters, acting as a data recorder platform, and becoming the means for decision-making regarding real-time robot behaviour. It is worth noting that we do not aim at implementing low-level control through the blockchain layer, owing to the lack of determinism and real-time capabilities in the proposed architecture, but instead drive high-level behaviour: paths to follow, state machines defining the mission flow or decisions in regard to how or when robots are meant to cooperate.

\begin{figure*}
    \centering
    \includegraphics[width=1.0\textwidth]{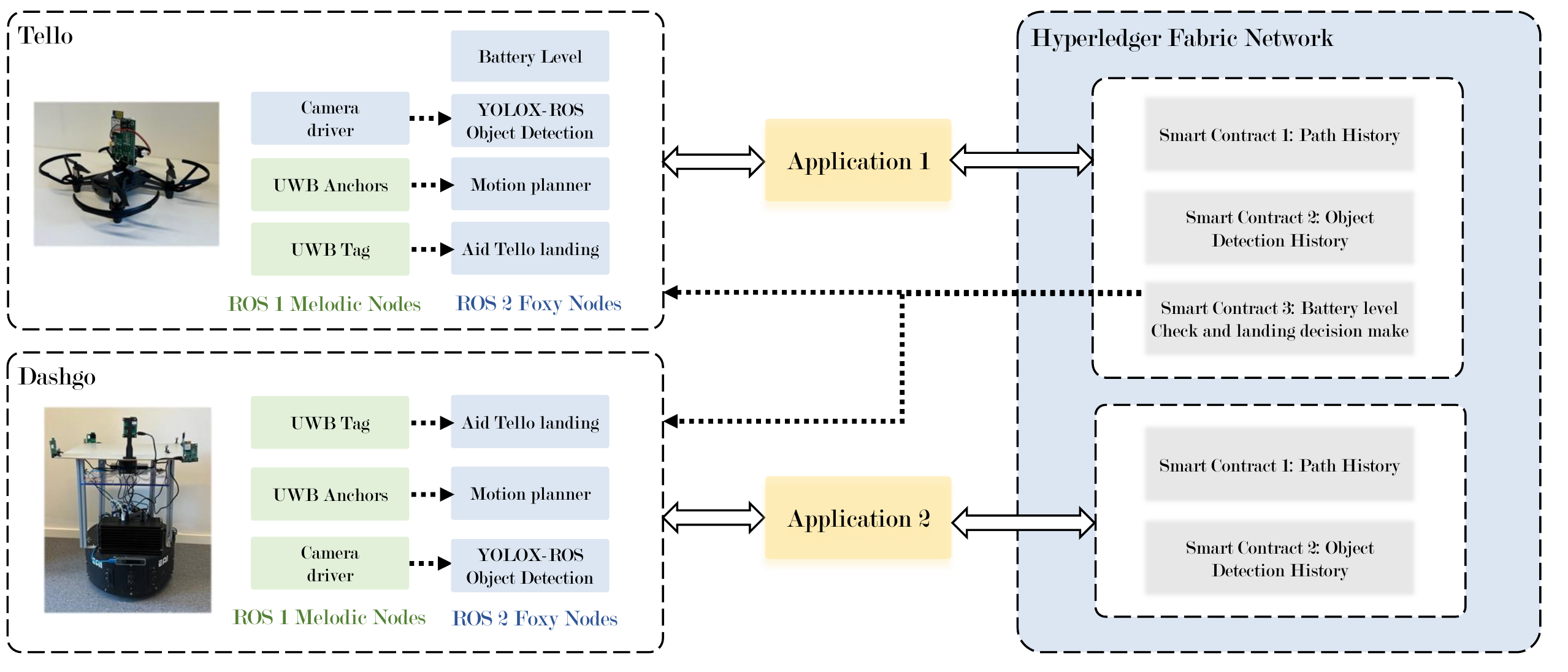}
    \caption{System implementation diagram.}
    \label{fig:implementation}
\end{figure*}

\section{Methodology}

This section describes the robots, hardware and software utilized in the experiments.

\subsection{Experimental platforms}

The experimental platform of this paper is consist of a commercially available Ryze Tello MAV and a ground robot. The drone is equipped with an UWB module for localization. We also utilize the drone camera for object detection, with the camera feed being available through a websocket connection to a controller computer. The ground robot is an EAI Dashgo platform equipped with an UP HD camera with an OV2735 sensor. An AAEON Up Xtreme with an Intel i7-8665UE processor is used as a companion computer on the Dashgo and RealSense T265 cameras has been used for VIO-based egomotion estimation.

For this experiment, four UWB nodes, Decawave's DWM1001 modules with custom firmware, had been deployed for robot localization and also five extra UWB nodes had been used on the Dashgo platform for more accurate docking of the Tello.

\subsection{Software}

Dashgo is running ROS Melodic under Ubuntu 18.04 for the main driver. Localization and object detection are running in ROS 2 Foxy. Fig.~\ref{fig:implementation} shows the different software modules running in different nodes. The fabric applications running onboard the robots are connected to peers running on a separate computer in the network with the same Intel i7 processor. 

To forward the data from Melodic topics to Foxy topics, the $ros1\_bridge$ package is used under the same computer. Also, to obtain camera images of the dashgo at a frequency of 30\,Hz even though they are forwarded to the object detector at 5\,Hz, the $usb\_cam$ package has been used which is available in both Melodic and Foxy. The object detector used in the experiments is YOLOX\footnote{https://github.com/Ar-Ray-code/YOLOX-ROS}, and a selection of objects are part of the categories in the COCO dataset which are used for the purpose of the inventory management.

To implement the different part of the system, the Go programming language (golang) has been used whenever possible to increase the potential for integration between the different parts of the system includes the smart contracts, applications and also ROS~2 nodes. To manage data securely and control robots, a private Hyperledger Fabric network has been set up.

To set up the private network, one orderer and two organizations including one peer each and corresponding CAs have been created. Then after genesis block generation and certificates creations for each organizations, the docker container has been brought up. Channel genesis block has been created for each channel and then the peers of organizations have been joined to the channel. After packaging and installing the definition of the chaincodes on peers, they have been then approved for organizations. In the last step, the chaincode has been invoked after commitment of definitions.

\subsection{Smart contracts}

Five smart contracts have been implemented in the system: two for storing the path tracking of two robots, two for storing the location of the detected objects by both robots and one for updating the battery level of the Tello in the asset and landing decision making(see Fig.~\ref{fig:implementation}). One application containing different functions such as creating new assets, read, update and changing the assets has been used for each robot.

A rendezvous point is defined for drone docking. Dasgho and Tello start inspecting the environment following the predefined path given by the operator through DLT where robot paths history and detected objects are stored in. According to the battery level of the Tello, when the battery level decreases below a threshold , the smart contract sends a docking order to the robots. When the ground robot accepts the drone for landing and charging, the position of the rendezvous point is given to the robots by smart contract. After the robots movement to a rendezvous point, the UWB receivers on the Dashgo platform are enabled to have relative localization and accurate docking for Tello drone, so external anchors are not used anymore for localization of the Tello drone. 

\subsection{Metrics}

In this paper our focus is on measuring and recording the data while using the Hyperledger Fabric blockchain for multi-robot collaboration to check the robots path. We analyze the memory and CPU usage of YOLOX and Fabric during the experiment and also latency of committing the transactions in the blockchain. Finally, we report the distribution of transaction latency in the blockchain layer.
 

\begin{figure}[t]
    \centering
    \setlength{\figureheight}{0.35\textwidth}
    \setlength{\figurewidth}{.5\textwidth} 
    \scriptsize{\input{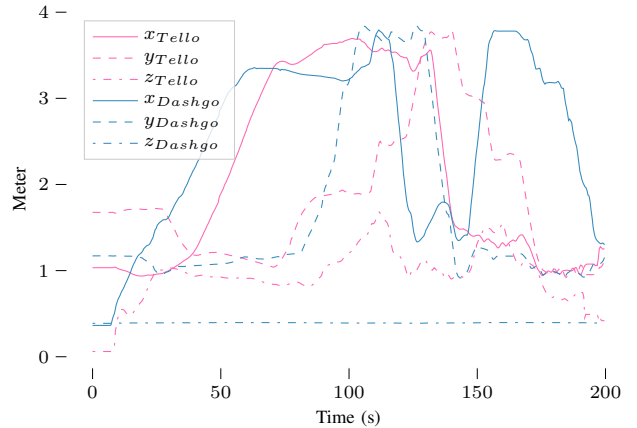}}
    \caption{Trajectory components for each of the robots. At the end of the mission, the robot trajectories converge as the Tello docks on top of the Dashgo.}
    \label{fig:trajectory}
    \vspace{-1.5em}
\end{figure}

\section{Experimental Results}

In this section we report the experimental results obtained by data from two robots which are following a predefined path for the purpose of detecting objects and storing them in a Fabric channel chaincode to manage the inventory. For this, UWB nodes, series of shelves and various objects from the COCO dataset categories are deployed in a $40\,m^2$ room. External UWB anchores are installed on the Dashgo platform, so that the Tello be able to have more accurate landing on Dashgo for charging.

\begin{figure}
    \begin{subfigure}[t]{0.48\textwidth}
        \centering
        \setlength{\figureheight}{.65\textwidth}
        \setlength{\figurewidth}{\textwidth} 
        \scriptsize{\input{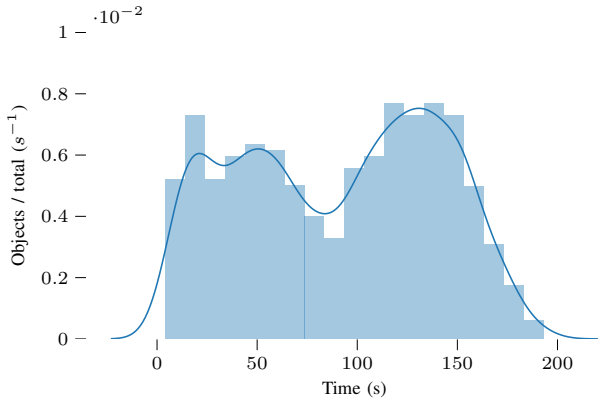}
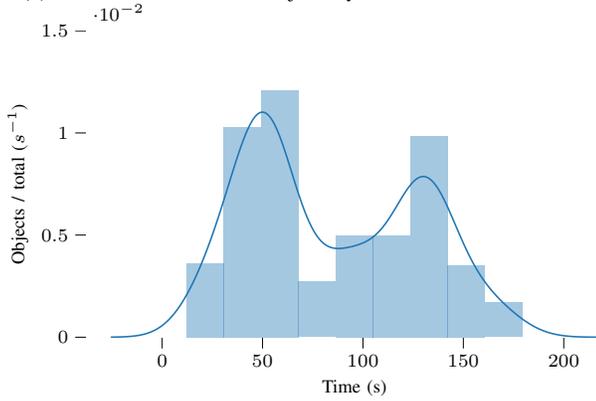}
        \caption{Distribution of detected objects by Tello over the mission time.}
        \label{fig:tello_distribution}
    \end{subfigure}
    \vspace{2em}
    \begin{subfigure}[t]{0.48\textwidth}
        \centering
        \setlength{\figureheight}{.65\textwidth}
        \setlength{\figurewidth}{\textwidth} 
        \scriptsize{
\begin{tikzpicture}

\definecolor{darkgray176}{RGB}{176,176,176}
\definecolor{lightgray204}{RGB}{204,204,204}
\definecolor{steelblue31119180}{RGB}{31,119,180}

\begin{axis}[
    height=\figureheight,
    width=\figurewidth,
    axis line style={white},
    legend style={fill opacity=0.8, draw opacity=1, text opacity=1, draw=lightgray204},
    tick align=outside,
    tick pos=left,
    x grid style={darkgray176},
    xlabel={Time (s)},
    ylabel={Objects / total ($s^{-1}$)},
    xmin=-37.6800999388089, xmax=229.160416938809,
    xtick style={color=black},
    y grid style={darkgray176},
    ymin=0, ymax=0.015,
    ytick style={color=black}
]
\draw[draw=none,fill=steelblue31119180,fill opacity=0.4] (axis cs:12.008878,0) rectangle (axis cs:30.6158292222222,0.00362060502271719);
\draw[draw=none,fill=steelblue31119180,fill opacity=0.4] (axis cs:30.6158292222222,0) rectangle (axis cs:49.2227804444444,0.010296095533352);
\draw[draw=none,fill=steelblue31119180,fill opacity=0.4] (axis cs:49.2227804444444,0) rectangle (axis cs:67.8297316666667,0.0121063980447106);
\draw[draw=none,fill=steelblue31119180,fill opacity=0.4] (axis cs:67.8297316666667,0) rectangle (axis cs:86.4366828888889,0.00271545376703789);
\draw[draw=none,fill=steelblue31119180,fill opacity=0.4] (axis cs:86.4366828888889,0) rectangle (axis cs:105.043634111111,0.00497833190623614);
\draw[draw=none,fill=steelblue31119180,fill opacity=0.4] (axis cs:105.043634111111,0) rectangle (axis cs:123.650585333333,0.00497833190623614);
\draw[draw=none,fill=steelblue31119180,fill opacity=0.4] (axis cs:123.650585333333,0) rectangle (axis cs:142.257536555556,0.00984351990551236);
\draw[draw=none,fill=steelblue31119180,fill opacity=0.4] (axis cs:142.257536555556,0) rectangle (axis cs:160.864487777778,0.00350746111575728);
\draw[draw=none,fill=steelblue31119180,fill opacity=0.4] (axis cs:160.864487777778,0) rectangle (axis cs:179.471439,0.00169715860439868);
\addplot [semithick, steelblue31119180, forget plot]
table {%
-25.5509855352808 4.02099094079007e-06
-24.3319790625644 5.5177711445497e-06
-23.112972589848 7.50803025681033e-06
-21.8939661171316 1.01307375462846e-05
-20.6749596444152 1.35559880208943e-05
-19.4559531716989 1.79895442550132e-05
-18.2369466989825 2.36773672875468e-05
-17.0179402262661 3.09099298926364e-05
-15.7989337535497 4.00260662975632e-05
-14.5799272808333 5.14160805659677e-05
-13.3609208081169 6.55238158539554e-05
-12.1419143354005 8.28473830902175e-05
-10.9229078626841 0.000103938264416661
-9.70390138996773 0.000129398547104365
-8.48489491725135 0.000159876109314645
-7.26588844453495 0.000196057669738103
-6.04688197181856 0.000238659726223108
-4.82787549910217 0.000288417538890371
-3.60886902638578 0.000346072453337975
-2.3898625536694 0.000412357999622308
-1.170856080953 0.000487985331423773
0.0481503917633859 0.000573628675088898
1.26715686447978 0.000669911528261033
2.48616333719617 0.000777394372117686
3.70516980991256 0.000896564631831051
4.92417628262895 0.00102782953218435
6.14318275534534 0.00117151234887509
7.36218922806173 0.00132785235502031
8.58119570077811 0.00149700851543084
9.80020217349451 0.00167906670126606
11.0192086462109 0.00187404990122805
12.2382151189273 0.00208193061166654
13.4572215916437 0.00230264431755955
14.6762280643601 0.00253610275038191
15.8952345370765 0.00278220544759856
17.1142410097928 0.00304084806012837
18.3332474825092 0.00331192587369059
19.5522539552256 0.00359533113836952
20.771260427942 0.0038909430437379
21.9902669006584 0.00419860953408049
23.2092733733748 0.00451812062231096
24.4282798460912 0.00484917341707207
25.6472863188076 0.00519132970217148
26.866292791524 0.00554396756978409
28.0852992642404 0.00590622927013327
29.3043057369567 0.0062769680559324
30.5233122096731 0.00665469732111031
31.7423186823895 0.00703754571081429
32.9613251551059 0.00742322206690771
34.1803316278223 0.00780899403095909
35.3993381005387 0.00819168382756533
36.6183445732551 0.0085676841830759
37.8373510459715 0.00893299650575947
39.0563575186879 0.00928329239153676
40.2753639914043 0.00961399827414578
41.4943704641206 0.00992040167895009
42.713376936837 0.0101977761501065
43.9323834095534 0.0104415205959377
45.1513898822698 0.0106473076343695
46.3703963549862 0.0108112346112488
47.5894028277026 0.010929970387972
48.808409300419 0.0110008908090783
50.0274157731354 0.0110221959965032
51.2464222458518 0.0109930032756837
52.4654287185682 0.010913410588549
53.6844351912845 0.0107845266284382
54.9034416640009 0.0106084655553678
56.1224481367173 0.0103883059113367
57.3414546094337 0.0101280151389805
58.5604610821501 0.00983234279693095
59.7794675548665 0.00950668705490972
60.9984740275829 0.00915694025183833
62.2174805002993 0.00878932014646492
63.4364869730157 0.00841019394623136
64.6554934457321 0.00802590226036717
65.8744999184484 0.00764258981000273
67.0935063911648 0.00726604908832002
68.3125128638812 0.00690158226260133
69.5315193365976 0.0065538855237741
70.750525809314 0.00622695889752507
71.9695322820304 0.00592404331070232
73.1885387547468 0.00564758552486005
74.4075452274632 0.00539923045965272
75.6265517001796 0.00517983947146188
76.845558172896 0.0049895323509641
78.0645646456124 0.00482775016728668
79.2835711183287 0.00469333561457646
80.5025775910451 0.00458462719962891
81.7215840637615 0.00449956343204051
82.9405905364779 0.00443579312428097
84.1595970091943 0.00439078796121387
85.3786034819107 0.00436195364140986
86.5976099546271 0.00434673611270005
87.8166164273435 0.00434271971067149
89.0356229000599 0.00434771435206606
90.2546293727763 0.00435982932769966
91.4736358454927 0.00437753167497618
92.692642318209 0.0043996875822945
93.9116487909254 0.00442558578075364
95.1306552636418 0.00445494240646498
96.3496617363582 0.00448788736283462
97.5686682090746 0.004524932768754
98.787674681791 0.00456692463666009
100.006681154507 0.00461497947288066
101.225687627224 0.00467040801819532
102.44469409994 0.00473462883315617
103.663700572657 0.00480907486197052
104.882707045373 0.0048950964602222
106.101713518089 0.00499386462400094
107.320719990806 0.0051062782902278
108.539726463522 0.00523287957156538
109.758732936238 0.00537378063013446
110.977739408955 0.00552860557459035
112.196745881671 0.00569645028536432
113.415752354388 0.00587586244284938
114.634758827104 0.00606484327267077
115.85376529982 0.00626087166010373
117.072771772537 0.00646095035960259
118.291778245253 0.00666167307886848
119.51078471797 0.00685931029691101
120.729791190686 0.00704991082942348
121.948797663402 0.00722941542677624
123.167804136119 0.00739377811829623
124.386810608835 0.0075390906309501
125.605817081552 0.00766170503046692
126.824823554268 0.00775834976650144
128.043830026984 0.00782623454765
129.262836499701 0.00786313991367128
130.481842972417 0.00786748798899445
131.700849445134 0.00783839166442597
132.91985591785 0.0077756803288462
134.138862390566 0.00767990122244118
135.357868863283 0.00755229646888944
136.576875335999 0.00739475682672768
137.795881808715 0.00720975414105379
139.014888281432 0.00700025533801093
140.233894754148 0.00676962155001648
141.452901226865 0.00652149655586809
142.671907699581 0.00625968913688014
143.890914172297 0.00598805416376409
145.109920645014 0.00571037722231045
146.32892711773 0.00543026735206576
147.547933590447 0.00515106201593431
148.766940063163 0.00487574775801282
149.985946535879 0.00460689917373732
151.204953008596 0.00434663785522936
152.423959481312 0.00409661194088546
153.642965954029 0.00385799585414175
154.861972426745 0.00363150882673868
156.080978899461 0.00341744992855065
157.299985372178 0.00321574662269839
158.518991844894 0.00302601337175586
159.73799831761 0.00284761656232199
160.957004790327 0.00267974199643294
162.176011263043 0.00252146140602252
163.39501773576 0.00237179485088399
164.614024208476 0.0022297664179914
165.833030681192 0.00209445129853183
167.052037153909 0.00196501302304905
168.271043626625 0.00184073033093673
169.490050099342 0.00172101379077683
170.709056572058 0.00160541283528708
171.928063044774 0.00149361430365742
173.147069517491 0.00138543388241563
174.366075990207 0.00128080200344213
175.585082462924 0.00117974580449388
176.80408893564 0.00108236870149925
178.023095408356 0.000988828985637812
179.242101881073 0.000899318666344803
180.461108353789 0.000814043557633229
181.680114826506 0.000733205370491213
182.899121299222 0.000656986345505542
184.118127771938 0.000585536749570082
185.337134244655 0.000518965376286223
186.556140717371 0.000457333035179068
187.775147190087 0.00040064889072988
188.994153662804 0.000348869416899117
190.21316013552 0.000301899663499385
191.432166608237 0.000259596484237283
192.651173080953 0.000221773349329052
193.870179553669 0.000188206355544418
195.089186026386 0.000158641051046377
196.308192499102 0.000132799709553788
197.527198971819 0.000110388716447785
198.746205444535 9.11057667741957e-05
199.965211917251 7.46466198165837e-05
201.184218389968 6.0711204947665e-05
202.403224862684 4.90089264953056e-05
203.622231335401 3.92630689169833e-05
204.841237808117 3.12142551913707e-05
206.060244280833 2.4622958706737e-05
207.27925075355 1.92711100996119e-05
208.498257226266 1.49628740128066e-05
209.717263698982 1.15246957262002e-05
210.936270171699 8.80473381277978e-06
212.155276644415 6.67180272150905e-06
213.374283117132 5.0139493272577e-06
214.593289589848 3.73678123667501e-06
215.812296062564 2.76165345429396e-06
217.031302535281 2.02380544151631e-06
};
\end{axis}

\end{tikzpicture}}
        \caption{Distribution of detected objects by Dashgo over the mission time.}
        \label{fig:dashgo_ditribution}
    \end{subfigure}
    \vspace{-1em}
    \caption{Figure (a) shows the distribution of detected objects by Tello, and (b) shows the distribution of detected objects by Dashgo.}
\end{figure}

Figure~\ref{fig:trajectory} shows the trajectories of the robots during the experiment. Towards the mission end, as it can be seen after more than three minutes, the positions converge when the Tello drone lands on the Dashgo. Figures~\ref{fig:tello_distribution} and~\ref{fig:dashgo_ditribution} show the distribution in time of the objects detected by Tello and Dashgo, respectively.

\begin{figure}
    \centering
    \setlength\figureheight{0.35\textwidth}
    \setlength\figurewidth{0.45\textwidth}
    \scriptsize{\input{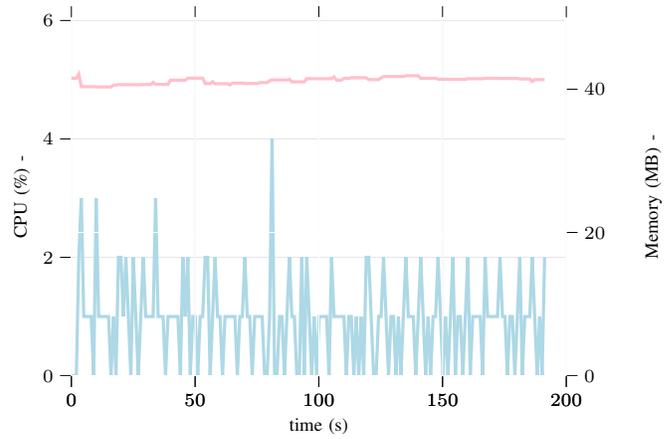}}
    \caption{Go applications activity during the mission where CPU usage is shown in blue and memory in pink.}
    \label{fig:application_activity}
\end{figure}

\begin{figure}
    \centering
    \setlength\figureheight{0.35\textwidth}
    \setlength\figurewidth{0.45\textwidth}
    \scriptsize{\input{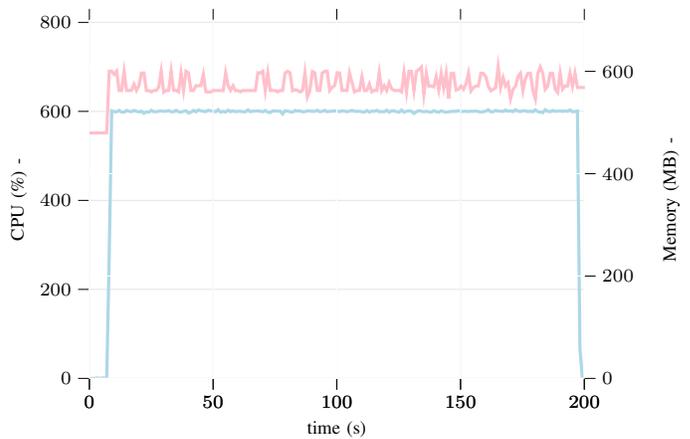}}
    \caption{YOLOX ROS 2 node activity during the mission where CPU usage is shown in blue and memory in pink.}
    \label{fig:yolox_activity}
\end{figure}

In Figure~\ref{fig:application_activity}, the CPU and memory utilization of the system during the experiments can be seen, while Figure~\ref{fig:yolox_activity} shows YOLOX resources utilization. We examined how integrating the Fabric network could affect an existing ROS~2 system, in terms of computational resources. Taking these results into account, we can conclude that the addition of Fabric as an additional data sharing channel is negligible, and that the proposed framework has the potential to be implemented on a variety of robotic platforms and application scenarios. 

\begin{figure}[tb]
    \centering
    \setlength\figureheight{0.42\textwidth}
    \setlength\figurewidth{0.48\textwidth}
    \scriptsize{
\begin{tikzpicture}

\definecolor{darkgray176}{RGB}{176,176,176}
\definecolor{darkorange25512714}{RGB}{255,127,14}

\definecolor{color4}{RGB}{255,255,204} 
\definecolor{color3}{RGB}{204,255,204} 
\definecolor{color2}{RGB}{204,255,255} 
\definecolor{color1}{RGB}{204,204,255} 
\definecolor{color0}{rgb}{1,0.752941176470588,0.796078431372549} 

\definecolor{color0}{rgb}{0.9,0.9,0.9}
\definecolor{color1}{rgb}{0.9,0.9,0.9}
\definecolor{color2}{rgb}{0.9,0.9,0.9}
\definecolor{color3}{rgb}{0.9,0.9,0.9}
\definecolor{color4}{rgb}{0.9,0.9,0.9}
\definecolor{color5}{rgb}{0.9,0.9,0.9}
\definecolor{color6}{rgb}{0.9,0.9,0.9}


\begin{axis}[
height=\figureheight,
width=\figurewidth,
axis line style={white},
legend style={fill opacity=0.8, draw opacity=1, text opacity=1, 
draw=white!80!black},
tick align=outside,
tick pos=left,
x grid style={white!69.0196078431373!black},
xmin=0.5, xmax=5.5,
xtick style={color=black},
xtick={1,2,3,4,5},
xticklabels={a,b,c,d,e},
y grid style={white!90!black},
ymajorgrids,
ylabel={Time (ms)},
ymin=-0.00894000000000001, ymax=1.2,
ytick style={color=black}
]
\addplot [black, fill=color0]
table {%
0.75 0.21455
1.25 0.21455
1.25 0.41655
0.75 0.41655
0.75 0.21455
};
\addplot [black]
table {%
1 0.21455
1 0.056
};
\addplot [black]
table {%
1 0.41655
1 0.7096
};
\addplot [black]
table {%
0.875 0.056
1.125 0.056
};
\addplot [black]
table {%
0.875 0.7096
1.125 0.7096
};
\addplot [black, mark=o, mark size=1.23, mark options={solid,fill opacity=0}, only marks]
table {%
1 0.7501
1 1.0773
1 0.8718
1 0.7248
1 0.8734
1 0.9911
1 0.7947
};
\addplot [black, fill=color1]
table {%
1.75 0.215
2.25 0.215
2.25 0.4142
1.75 0.4142
1.75 0.215
};
\addplot [black]
table {%
2 0.215
2 0.0522
};
\addplot [black]
table {%
2 0.4142
2 0.7092
};
\addplot [black]
table {%
1.875 0.0522
2.125 0.0522
};
\addplot [black]
table {%
1.875 0.7092
2.125 0.7092
};
\addplot [black, mark=o, mark size=1.23, mark options={solid,fill opacity=0}, only marks]
table {%
2 1.0456
2 1.0526
2 0.7497
2 1.0703
2 0.8777
2 0.7302
2 0.876
2 0.9943
2 0.7962
};
\addplot [black, fill=color2]
table {%
2.75 0.0686
3.25 0.0686
3.25 0.2884
2.75 0.2884
2.75 0.0686
};
\addplot [black]
table {%
3 0.0686
3 0.0466
};
\addplot [black]
table {%
3 0.2884
3 0.6132
};
\addplot [black]
table {%
2.875 0.0466
3.125 0.0466
};
\addplot [black]
table {%
2.875 0.6132
3.125 0.6132
};
\addplot [black, mark=o, mark size=1.23, mark options={solid,fill opacity=0}, only marks]
table {%
3 0.6536
3 0.6536
3 0.7618
3 0.7199
3 0.6214
3 0.621
3 0.6236
3 0.6237
3 0.8834
3 0.8835
3 0.8835
3 0.8159
3 0.8154
3 0.6662
3 0.6664
3 0.7251
3 0.7252
3 0.74
3 0.6799
3 0.7638
3 0.8687
3 0.8688
3 0.6339
3 0.8637
3 0.7053
};
\addplot [black, fill=color3]
table {%
3.75 0.1304
4.25 0.1304
4.25 0.3651
3.75 0.3651
3.75 0.1304
};
\addplot [black]
table {%
4 0.1304
4 0.0505
};
\addplot [black]
table {%
4 0.3651
4 0.637
};
\addplot [black]
table {%
3.875 0.0505
4.125 0.0505
};
\addplot [black]
table {%
3.875 0.637
4.125 0.637
};
\addplot [black, mark=o, mark size=1.23, mark options={solid,fill opacity=0}, only marks]
table {%
4 0.8214
4 1.0525
4 0.8874
4 0.7241
};
\addplot [black, fill=color4]
table {%
4.75 0.1537
5.25 0.1537
5.25 0.748475
4.75 0.748475
4.75 0.1537
};
\addplot [black]
table {%
5 0.1537
5 0.0431
};
\addplot [black]
table {%
5 0.748475
5 1.0839
};
\addplot [black]
table {%
4.875 0.0431
5.125 0.0431
};
\addplot [black]
table {%
4.875 1.0839
5.125 1.0839
};
\addplot [black]
table {%
0.75 0.25985
1.25 0.25985
};
\addplot [black]
table {%
1.75 0.2599
2.25 0.2599
};
\addplot [black]
table {%
2.75 0.2117
3.25 0.2117
};
\addplot [black]
table {%
3.75 0.2179
4.25 0.2179
};
\addplot [black]
table {%
4.75 0.50075
5.25 0.50075
};
\end{axis}

\end{tikzpicture}}
    \caption{Distribution of the latency for committing transactions between the robot and the peer node where (a) shows the chaincode recording the Tello battery level, (b) shows the chaincode recording the Tello path history, (c) stores objects detected by the Tello drone, (d) shows the Dashgo path history recorder and (e) shows the storage of objects detected by Dashgo.}
    \label{fig:latency}
\end{figure}
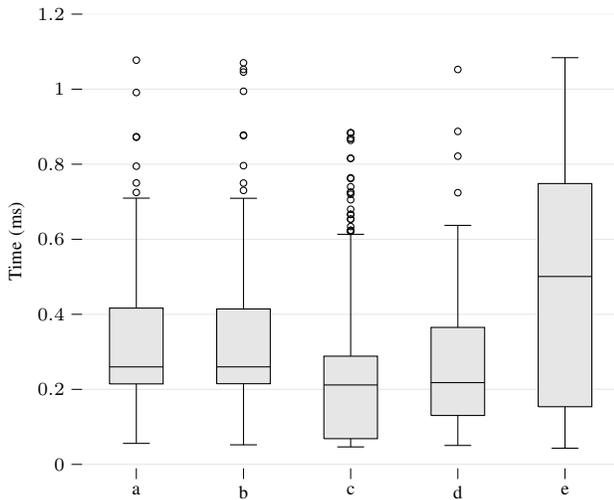

Asset creation is a key parameter affecting the system latency because each transaction is only confirmed when it appears in a block. Therefore, in addition to the resource utilization, we have measured the latency of storing data in the blockchain while application nodes are running on the robots and peers and orderer are running in another computer in the same network. Figure~\ref{fig:latency} shows the latency distribution of the five smart contracts where over 200\,HZ of ROS~2 data has been stored in average. The main parameters affecting the transaction commit latency are the minimum amount of transactions to be included in a block, and the maximum timeout for a transaction to be included. A block is thus created whenever one or another amount is reached. An initial study on the performance of Fabric with different parameters has already been shown in~\cite{salimi2022towards}. In this paper, we analyze the transaction latency distribution to illustrate its relation to the data types and frequencies in each chaincode. Path history and Tello battery are recorded at a constant, pre-defined frequency, and thus the distributions are more similar. The values are not constant because network delays or other transactions still affect the process. The Tello object detection chaincode has slightly lower latency owing to the higher frequency of the data over most of the mission (see Fig.~\ref{fig:tello_distribution}). At the same time, the Dashgo distribution is wider as the recording frequnecy is more irregular (see Fig.~\ref{fig:dashgo_ditribution}.

\section{Conclusion}\label{sec:conclusion}

In this paper, a framework leveraging blockchain technology for multi-robot collaboration in industrial applications for the purpose of identity management, data sharing security, monitoring and multi-robot inventory management and consensus has been presented. DLTs has been used in order to protect the data sharing in multi-robot collaboration, while providing samples of applications transfering data between ROS~2 nodes and smart contracts in Hyperledger Fabric. On the other hand, for having high performance and low system complexity, robots localization and landing have been done through UWB positioning. According to the experimental results, it has been shown that this framework provides great amount of security and identity management features, and at the same time adding the blockchain layer has minimal impact on the utilization of computational resources.


\section*{Acknowledgment}

This research work is supported by the Academy of Finland's RoboMesh and AeroPolis projects (grant numbers 336061 and 348480).

\bibliographystyle{unsrt}
\bibliography{bibliography}

\end{document}